\documentclass{article}

\usepackage[final]{neurips_2026}

\makeatletter
\gdef\@noticestring{Presented at the InterpScience Workshop at the 40th Conference on Neural Information Processing Systems (NeurIPS 2026).}

\makeatother
\usepackage[utf8]{inputenc} 
\usepackage[T1]{fontenc}    
\usepackage{hyperref}       
\usepackage{url}            
\usepackage{booktabs}       
\usepackage{amsmath}        
\usepackage{amsfonts}       
\usepackage{nicefrac}       
\usepackage{microtype}      
\usepackage{xcolor}         
\usepackage{natbib}         
\usepackage{graphicx}
\usepackage{float}
\usepackage{multirow}       

\title{Text Dominance in Paralinguistic Conflict: Tracing Suppression Through Audio LLM Circuits}

\author{%
  \textbf{Ankur Banerjee}$^{*, 1}$ \quad
  \textbf{Vishnu Guntakala}$^{*, 1}$ \quad
  \textbf{Danity Pike}$^{\dagger, 1}$ \\
  \textbf{Advay Roongta}$^{\dagger, 1}$ \quad
  \textbf{Kevin Yang}$^2$ \quad
  \textbf{Ruizhe Li}$^3$ \\[1em]
  $^1$Independent Researcher\qquad $^2$Brown University \qquad $^3$University of Birmingham \\
  \texttt{\{ankurbaner09, vishnun1231, roongtaadvay\}@gmail.com} \\
  \texttt{pikedm27@episcopalacademy.org} \quad \texttt{kevin\_c\_yang@brown.edu} \quad \texttt{r.li.7@bham.ac.uk}
}

\begin{document}

\maketitle
\renewcommand{\thefootnote}{\fnsymbol{footnote}}
\footnotetext[1]{Equal contribution.}
\footnotetext[2]{Equal contribution.}
\renewcommand{\thefootnote}{\arabic{footnote}}

\begin{abstract}
Audio LLMs deployed in settings such as crisis hotlines and customer support are
only useful beyond speech-to-text pipelines if they use acoustic cues rather than
the transcript alone. We test whether they prioritise transcript content over
paralinguistic cues (emotion, speaker identity, sarcasm) and trace the mechanism
when they do. Across six audio LLMs under LISTEN-style conflict conditions
\citep{chen2026listen}, the mean text-dominance rate on mismatched text+audio
items is 0.78 (0.86 excluding sarcasm). In the most text-dominant screened model,
probing shows speaker identity remains linearly decodable at ceiling at the
language model's input, so the failure is not perceptual. Back-patching, reported
to repair lexical text dominance, has no measurable effect here; but it also
fails on its own source task in our replication ($n{=}300$), so we treat its
null as uninformative rather than as evidence of non-transfer. Attention knockout partially restores
audio-grounded output by blocking the claim from propagating
through intermediate tokens during open-ended generation, whereas blocking the
answer position’s direct connection to the conflicting claim has no effect. The claim's influence therefore travels through an
intermediate relay rather than a single severable edge. The effect appears only
in free-form generation and not in closed-set multiple choice, suggesting that
benchmarks relying on that format may be systematically blind to interventions
of this kind. Code and data are publicly available at \url{https://github.com/AdvayRoongta/audio-llm-mech-interp}.
\end{abstract}

\section{Introduction}

Audio LLMs are increasingly used in conversational applications, including customer
service, interview transcription, and voice-based assistants \citep{chu2024qwen2audio,
tang2024salmonn}. But they are largely trained and evaluated on transcription and other
lexically-centred tasks, which encourages them to prioritise the words spoken over
acoustic and non-verbal information such as prosody, emotion, and speaker identity. That
tendency undercuts the central advantage of processing raw audio: access to information
an automatic speech recognition pipeline would discard before a text-based LLM ever saw
it.

The cost is concrete wherever voice carries meaning the transcript does not. A caller on
a crisis line may say everything is fine while their prosody says otherwise; a speaker's
voice may contradict their claimed identity, a failure mode central to fraud detection
\citep{hansen2015speaker}; a customer's sarcasm may exist only in their tone. Depression
and trauma screening depend on vocal affect for the same reason \citep{cummins2015review}.
In each case the non-verbal signal modifies or contradicts the lexical content, and a
text-dominant model discards it.

Text dominance in audio LLMs has been evaluated behaviourally, typically by pairing audio
with transcript text asserting a conflicting attribute \citep{chen2026listen, mcrbench},
and studied mechanistically for \emph{lexical} conflicts, where back-patching has been
proposed as a training-free repair \citep{cho2026conflict}. What remains unexamined is the
mechanism behind suppression of \emph{paralinguistic} evidence: whether the circuits that
produce lexical text dominance also govern conflicts carried by prosody, emotion, and
speaker identity, and whether the interventions developed there transfer. We localise the
responsible components, test those interventions, and identify one that works.

We approach this problem mechanistically rather than treating text dominance as a purely
behavioural failure. Observing that a model follows text over audio does not tell us why:
the model may have failed to encode the acoustic signal in the first place, or it may
have encoded the signal but failed to use it when resolving the conflict. We therefore
investigate the failure in four stages. First, linear probing tests whether
paralinguistic information survives the audio encoder and projector and remains available
to the language model. Second, attribution patching identifies where the model relies on
the audio and textual signals when they conflict. Third, back-patching and direct
amplification test whether the audio pathway can be strengthened enough to overcome the
textual signal. Finally, attention knockout tests the converse hypothesis: that reducing
the influence of the conflicting text can restore audio-grounded behaviour. We first establish whether the relevant information is present, then ask how it is used, and finally intervene on the pathway
implicated by that analysis. To avoid attributing missing information to suppression, we
apply this mechanistic procedure only to capabilities the model can recover from audio
alone.

Our contributions are fourfold. First, we introduce a 3,594-item paralinguistic conflict
benchmark spanning emotional prosody, speaker identity, and sarcasm/pacing, and use it to
measure text dominance across six audio-language models and two encoder families. Second,
we show that in the model selected for mechanistic analysis, paralinguistic information
remains fully recoverable at the language model's input, reaching ceiling-level probe
accuracy from encoder layer 13 onward; the observed text dominance is therefore not
explained by a failure to perceive or encode the acoustic signal. Third, we systematically
test whether the audio pathway can be strengthened through activation back-patching and
direct amplification, and find no configuration that restores audio-grounded answers;
replicating back-patching on the lexical task for which it was designed also yields no
repair, so the intervention provides no working positive control and its null on
paralinguistic conflict is uninformative about mechanism. Fourth, we localise the textual
influence to an intermediate relay pathway rather than a direct claim-to-answer
connection, and show that attenuating this pathway during free-form generation reduces
text dominance substantially while preserving matched-item and text-only performance.
This effect is not captured by the closed-set multiple-choice evaluation used in standard
audio-language-model benchmarks.

\section{Related Work}

\paragraph{Text dominance in audio LLMs.}
Audio LLMs encode raw audio and map that representation into the language model's input
space alongside the text instruction \citep{chu2024qwen2audio, tang2024salmonn,
radford2023whisper}, giving the model access to acoustic detail that transcription would
discard. In practice they follow textual information over contradictory audio, a
behaviour known as text dominance. \citet{cho2026conflict} analysed this mechanistically
and found that text and audio propagate through distinct pathways early in processing
before converging in later layers, with the text branch inhibiting rather than replacing
audio representations; they proposed back-patching as a training-free repair.
\citet{billa2026alme} examined audio--text conflicts across languages with the ALME
benchmark and found that audio--text conflicts produce far higher text-dominance rates
than text--text conflicts. Both address conflicts in lexical or semantic content, whereas
we ask whether the same mechanisms apply when the conflicting information is carried by
prosody, emotion, or speaker identity.

\paragraph{Mechanistic interpretability methods.}
Mechanistic interpretability identifies the internal components involved in a model's
behaviour rather than characterising that behaviour only at the level of inputs and
outputs \citep{wang2023ioi}. Activation patching is the standard causal tool: a
component's activation is replaced with one from a clean or corrupted input and the change
in output is measured \citep{vig2020causal, meng2022rome}. Done exhaustively it requires
one forward pass per component, so attribution patching substitutes a gradient-based
approximation requiring far fewer \citep{syed2023attribution}. That approximation can
misrank components, so a circuit analysis must include a faithfulness check verifying that
the selected components reproduce the full model's behaviour \citep{wang2023ioi,
hanna2024faithfulness}. \citet{cho2026conflict} applied this pipeline to lexical text
dominance and proposed back-patching, in which late-layer activations are routed into
early layers. Whether the same circuit or the same repair applies to paralinguistic
conflict has not been tested.

\paragraph{Paralinguistic representation and benchmarks.}
Paralinguistic cues include tone, pitch, intensity, timing, emphasis, and speaker
characteristics such as age or gender \citep{schuller2013paralinguistics}, none of which
are recoverable from a transcript alone. \citet{chen2026listen} introduced the LISTEN
benchmark, separating lexical from acoustic information through neutral, matched,
mismatched, and paralinguistic conditions, and found consistent lexical dominance across
six audio LLMs, with models often defaulting to neutral interpretations under conflict.
MCR-Bench \citep{mcrbench} measures adherence to adversarial textual descriptions on sound
and vocal classification, and PARSA-Bench \citep{parsabench} reports that text-only
baselines outperform audio counterparts in Persian, indicating the reliance is not
confined to English. A parallel line of work steers attention at inference time to resolve
conflicts between competing \emph{textual} sources: PASTA \citep{zhang2024pasta} reweights
a small set of attention heads to emphasise user-specified spans, while other methods
shift activations along probe-identified directions \citep{li2023iti} or use contrastive
decoding that upweights context-conditioned predictions \citep{shi2024cad}; the broader
problem of contextual versus parametric knowledge conflict has been studied extensively in
text-only settings \citep{longpre2021entity, xu2024knowledgeconflict}. These establish
behavioural rates or operate entirely within a single textual stream; none localises the
components responsible when the competing evidence is acoustic and must first pass through
an audio encoder and projector. Appendix~\ref{app:related} gives an extended treatment.

\section{Method}

\begin{figure}[t]
  \centering
  \includegraphics[width=\linewidth]{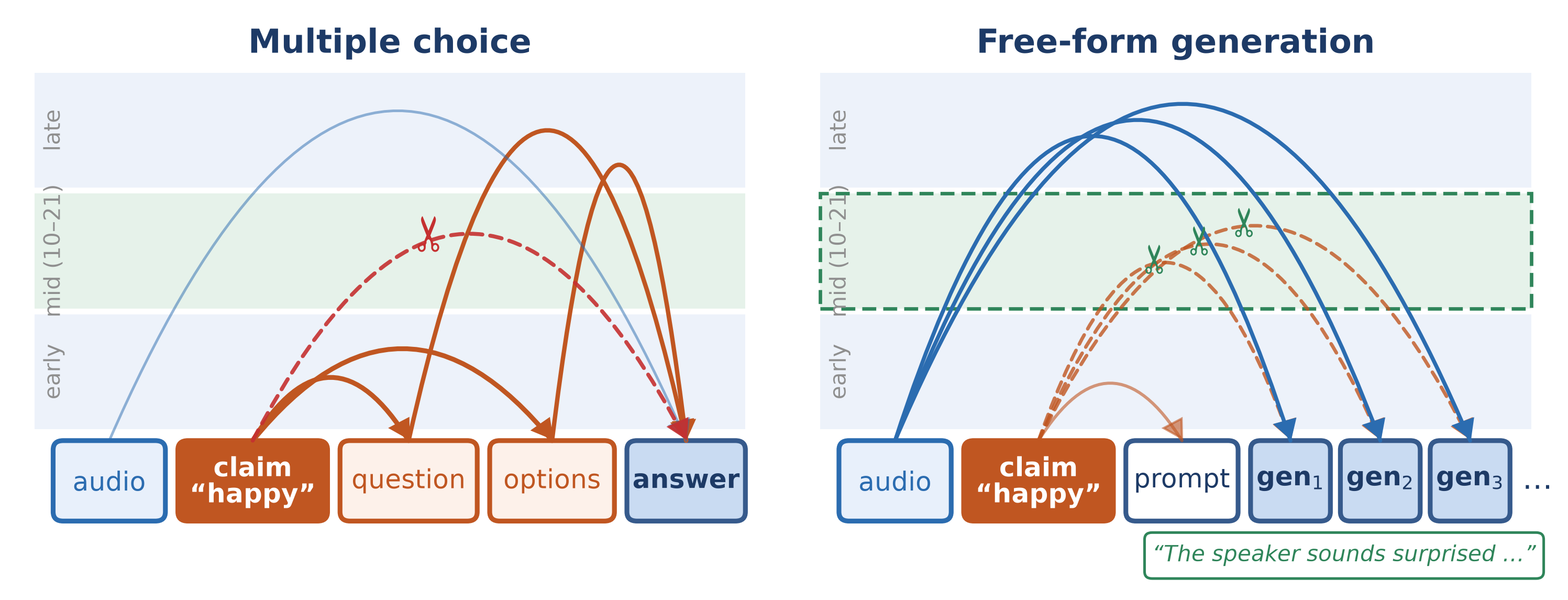}
  \caption{Attention knockout, the intervention at the centre of our analysis.
Tokens run left to right; each edge is one position attending to another, drawn
at the layer band where that attention occurs. Left: in closed-set multiple
choice the claim is copied into the question and option tokens early and the
answer position reads it from them, so blocking the direct claim$\to$answer edge
at all 32 layers leaves TDR unchanged (0.98 $\to$ 0.98). Right: in free-form
generation we block every generated token from the claim span at mid layers
(10--21), re-applied at each decoding step; with the relay cut, generation
follows the audio (gen-TDR 1.00 $\to$ 0.38). The direct-edge and mid-layer
blocks are the measured interventions; the placement of relay-building and
audio-reading edges in the early and late bands is schematic.}
  \label{fig:method}
\end{figure}

\subsection{Dataset Construction and Model Selection}

We created a 3,594 item dataset in three categories from existing speech corpora with acoustic and prosodic annotations, using labels provided by MUStARD++ \citep{ray2022mustardpp} and emotion information encoded in RAVDESS \citep{livingstone2018ravdess} file names. Using the LISTEN benchmark \citep{chen2026listen}, we created 3 types of conflicts: Emotional Prosody Conflict (e.g., cheerful text paired with an angry vocal tone), Speaker Identity Conflict (e.g., a transcript claiming a male speaker paired with a female voice), and Sarcasm Pacing Conflict (e.g., a sincere transcript paired with an overtly sarcastic tone).

We evaluated 6 audio language models using two families of audio encoders to determine whether text dominance is a phenomenon specific to the Whisper-based encoders \citep{radford2023whisper} or common across different encoder architectures. Four models utilize Whisper-family encoders: Qwen2-Audio \citep{chu2024qwen2audio}, 

\citep{tang2024salmonn}, DeSTA2.5 \citep{lu2025desta25}, and Audex \citep{nvidia2026audex}, a 30-billion-parameter model included to test whether text dominance continues to occur at a larger scale. Two models use non-Whisper audio encoders: Phi-4-multimodal \citep{abouelenin2025phi4}, which uses a custom convolutional-conformer encoder, and Baichuan-Omni-1.5 \citep{li2025baichuan}, which uses a residual-vector quantized audio tokenizer. Baichuan-Omni-1.5 shares the Qwen2.5-7B \citep{qwen2025qwen25} language-model backbone with Qwen2-Audio, but we include it because our comparison concerns the audio encoder rather than the language model backbone. All inferences were run with 16-bit precision to prevent confoundation of accuracy and text-dominance measurements central to our analysis.

\subsection{Mechanistic Interpretability Procedure}

Figure~\ref{fig:method} summarizes the procedure. To identify the mechanism behind this behavior in the model showing the strongest bias, we probed internal activations, applied attribution patching, tested back-patching, and used attention knockout. In the first set of experiments, we applied linear probing \citep{alain2016probing, belinkov2022probing} on the activation of the model layers without updating their weights to verify whether paralinguistic cues are encoded at each level of representation starting from the audio encoder up to the projector layer where the language model receives input. Second, attribution patching \citep{syed2023attribution, hanna2024faithfulness} was performed as a gradient-based approximation approach to estimate the causal effect of each component on the model's decisions without the cost of testing them independently, and confirmed its faithfulness by verifying if the highly weighted components can replicate the output of the model alone. Third, back-patching \citep{cho2026conflict} was performed as an intervention from previous research related to lexical text dominance, where activations from the later layers were replicated into earlier layers varying on multiple dimensions to determine whether it transfers to paralinguistic conflict. Fourth, we used attention knockout \citep{geva2023dissecting} to examine how the effect of the claim would propagate within the neural network by knocking out the attention of specific token positions to the false text claim. We evaluated each intervention under both multiple-choice and open-ended generation settings. Isolated controls determined if any effect seen was because of actual repair versus damage done to the output of the model.

\section{Results}

\subsection{Statistical Significance of Text-Dominance Differences}

\begin{figure}[!htbp]
  \centering
  \includegraphics[width=0.85\linewidth]{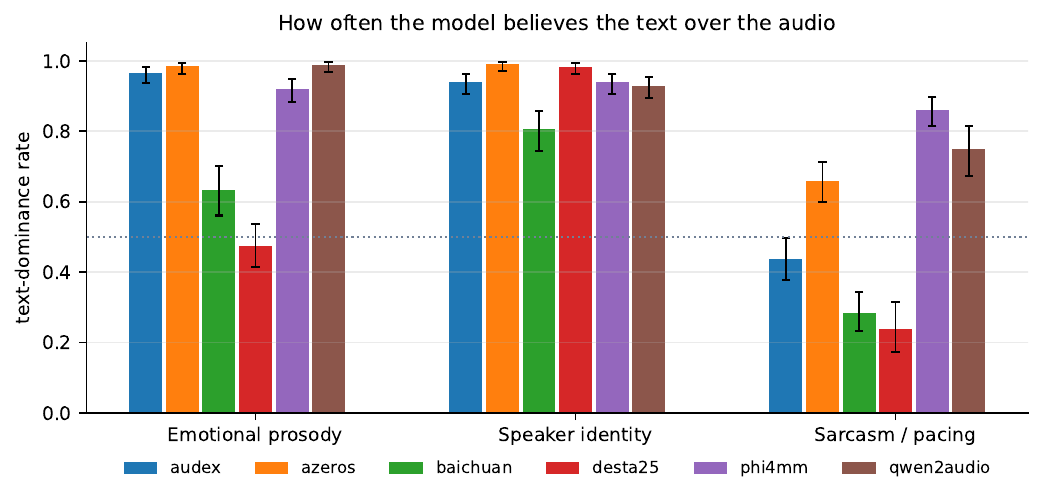}
  \caption{Text-dominance rate on mismatched \texttt{text\_audio} items, by model
  and conflict category. Error bars are 95\% Wilson intervals; the dotted line
  marks chance. SALMONN-13B is omitted because its behavioural run predates
  modality-split scoring; its text-dominance rates appear in
  Table~\ref{tab:app-tdr}. AZeroS was evaluated behaviourally and is shown
  for completeness; it is not among the six models in the statistical
  comparison.}
  \label{fig:tdr-model}
\end{figure}

Text-dominance rates differ significantly between models in all three categories
(Cochran's Q \citep{cochran1950}, the appropriate omnibus test for binary
outcomes measured repeatedly on the same items: emotional prosody $Q=201.04$,
$df=5$, $n=304$; speaker identity $Q=218.09$, $df=5$, $n=331$; sarcasm/pacing
$Q=345.00$, $df=5$, $n=369$; all $p<10^{-40}$). Complete-case $n$ varies because
we drop items on which any model returned an unparseable response, which
Baichuan-Omni-1.5 accounts for almost entirely. Averaged over the six models above on cells with modality-split scoring, the text-dominance rate on mismatched text+audio items is 0.78 (0.86 excluding sarcasm); SALMONN is omitted because its run predates modality-split scoring, and pooling in text-only items would inflate the rate (Figure~\ref{fig:tdr-model}).

Pairwise McNemar tests \citep{mcnemar1947} with Holm--Bonferroni correction
\citep{holm1979} across all 15 comparisons are reported in
Appendix~\ref{app:pairwise} (Tables~\ref{tab:app-mcnemar-prosody}--\ref{tab:app-mcnemar-sarcasm}).
We note one result here because it bounds what our behavioural data can support:
on speaker identity, the five models other than SALMONN are statistically
indistinguishable from one another despite an 8-point spread in raw
text-dominance rate, so that spread should not be read as a model ranking at
this sample size.

Our mechanistic findings come from a single model. Qwen2-Audio-7B-Instruct
\citep{chu2024qwen2audio} showed the strongest text-dominance bias among models
passing capability screening, and all probing, attribution patching,
back-patching and attention-knockout results below are from it.

\subsection{Capability Screening}

Attribution-based localization assumes that the model can perform the task on
clean inputs: where audio-only accuracy is at chance, a high text-dominance rate
reflects the absence of a usable acoustic representation rather than the
suppression of one \citep{chen2026listen}. We therefore screen every model in
every category on the \texttt{Paralinguistic} (audio-only) condition and keep only those exceeding chance for mechanistic analysis.

Five of six models pass the screen in both categories we analyze mechanistically.
Baichuan-Omni-1.5 falls below chance on emotional prosody (0.099 against 0.125)
and is excluded from that category. SALMONN-13B clears emotional prosody (0.150
against 0.125) but falls below chance on speaker identity (0.453 against 0.500)
under the option log-probability scoring used by the mechanistic screen, and is
excluded on that basis.

We note that SALMONN's behavioural accuracy on speaker identity (0.600) exceeds
its screened accuracy, because the behavioural scorer excludes responses falling
outside the option set from its denominator and SALMONN produces such responses
on roughly a fifth of items. The log-probability screen cannot fail to parse and
is the criterion we apply throughout. The screen therefore measures whether a
model answers correctly from audio in our prompt format, and cannot separate a
model that does not perceive prosody from one that perceives it but does not
answer in the form we ask; we treat the exclusion as a statement about what we
can measure rather than about what SALMONN-13B can hear.

All probing, attribution patching, back-patching and attention-knockout results in
the main text are from Qwen2-Audio-7B-Instruct \citep{chu2024qwen2audio}, which
shows the strongest and most consistently measurable bias among screened models.
It uses a Whisper-family encoder \citep{radford2023whisper}, so our account may be
specific to that encoder family; the two non-Whisper models in our behavioural
set would settle this, and we did not run them mechanistically.

\subsection{Audio Information Retention and Attribution}
\label{sec:retention}

\begin{figure}[!htbp]
  \centering
  \includegraphics[width=0.8\linewidth]{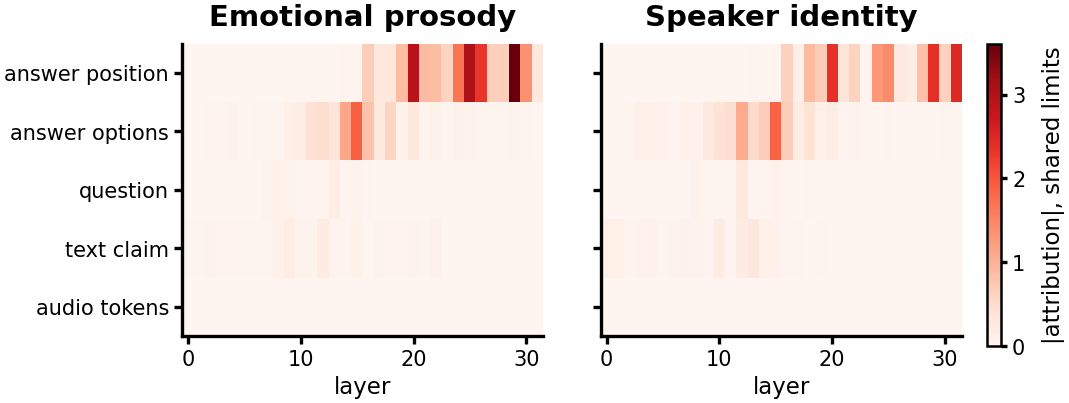}
  \caption{Qwen2-Audio-7B text circuit attribution by layer and prompt position}
  \label{fig:layers-acc}
\end{figure}

Next, we evaluate whether audio information is retained in the model prior to the decoder and reasoning steps. Using a linear probe classifier \citep{alain2016probing}, we found that speaker gender was recovered at an accuracy of 1.000 in the encoder from layer 13 onward, and that this information was not lost in the projector. This is consistent with prior benchmark findings that speaker gender is near-ceiling for models in this family \citep{parsabench}. We paired this with attribution patching \citep{syed2023attribution} to identify how much audio information was truly used by the model. Text was found to contribute 3.29 compared to audio's 0.18 contribution, a ratio of roughly 18:1. We applied a faithfulness test \citep{wang2023ioi, hanna2024faithfulness} by comparing the output of the full model to the top 20\% of components identified by attribution patching, which reproduced 100\% of the original model's behavior.

\subsection{Backpatching Produces No Measurable Repair}

Due to the disproportionate signal in text, we tried backpatching which previous work has shown reduces TDR for lexical text dominance \citep{cho2026conflict}. The hidden states of the audio token span in decoder layers 31/32, identified by their position within the sequence, were written into earlier layer 0, and the model was re-run from that point. We expected this to work because adding a later activation into an earlier one would give the model another opportunity to process the audio evidence, increasing its effective weight.

Attribution data from the previous experiment revealed the influence of audio peaks at layer 9 and falls to 0.5\% by the later layers. The best source layer was layer 16, which gained 0.511 nats against a baseline of $-7.32$ nats, closing 7\% of the gap; sources in later layers made things worse, with margins falling as low as $-0.381$ nats below baseline. Widening the destination window (widths 1, 3, 5, and 7) proved ineffective, as did amplifying the audio span (1.5x, 2.0x, 3.0x, 5.0x). At the highest amplification (5.0x), capability was reduced from 1.000 to 0.100, while the decision margin became undefined. The norm profile grew 4.9x between the early and late layers, but rescaling the injected vectors to compensate did not help either. 

To establish whether this intervention functions as a positive control in our pipeline, we additionally rebuilt the lexical conflict task it was developed for from the ALME benchmark \citep{billa2026alme}, resynthesizing audio with TTS, and ran each of the four English flip types at its own published layer configuration. Results are given in Table~\ref{tab:alme}.

The phenomenon reproduces in kind --- text dominance of 0.793 against an audio-only ceiling of 0.908 --- but the repair does not: back-patching yields no measurable gain on the lexical task either, altering the outcome on 26 of 300 items and helping on 14 while hurting on 12. Because the published configurations are tuned per flip type on the original authors' audio and we resynthesized with a different TTS system, we read this as configuration sensitivity of a layer-pinned intervention rather than as a challenge to the original result. What these results jointly establish is narrower than transfer failure: back-patching provides no working positive control in our hands, in either setting, so we draw no conclusion from its null about the relationship between lexical and paralinguistic mechanisms. The audio-side evidence that does bear weight comes from probing and amplification, reported above and taken up in Section~\ref{sec:discussion}.

\begin{table}[!htbp]
\centering
\caption{Replication of back-patching on the lexical conflict task it was
developed for, rebuilt from ALME \citep{billa2026alme} with TTS resynthesis.
Each flip type is run at its own published layer configuration, pinned rather
than swept. Paired 95\% CI and McNemar exact test on discordant pairs.}
\label{tab:alme}
\small
\begin{tabular}{lcccc}
\toprule
Flip type & Config (src$\rightarrow$dst, window) & $n$ & Baseline $\rightarrow$ Patched & Gain \\
\midrule
Adjective swap & $15\rightarrow4$, $w{=}3$ & 75 & 0.213 $\rightarrow$ 0.227 & $+0.013$ \\
Number swap    & $12\rightarrow4$, $w{=}1$ & 75 & 0.213 $\rightarrow$ 0.213 & $+0.000$ \\
Time swap      & $13\rightarrow4$, $w{=}3$ & 75 & 0.320 $\rightarrow$ 0.360 & $+0.040$ \\
Negation add   & $31\rightarrow6$, $w{=}3$ & 75 & 0.080 $\rightarrow$ 0.053 & $-0.027$ \\
\midrule
\textbf{Pooled} & --- & \textbf{300} & \textbf{0.207 $\rightarrow$ 0.213} & \textbf{$+0.007$} \\
\multicolumn{5}{l}{\footnotesize Paired 95\% CI $[-0.027, +0.040]$; 14 items fixed, 12 broken; McNemar exact $p = 0.845$.} \\
\multicolumn{5}{l}{\footnotesize Reported in prior work: $+0.160$ \citep{cho2026conflict}.} \\
\bottomrule
\end{tabular}
\end{table}

\subsection{Attention Knockout}

We next tried reducing the text's weight instead of increasing the audio's. We applied attention knockout \citep{geva2023dissecting}, blocking the answer position from attending to the text claim tokens, to prevent the claim from carrying too much weight when audio evidence contradicted it. This did not reduce TDR because information from the counterfactual text claim was merged into the question and option tokens early: by the time the model answers, the claim is no longer confined to the claim tokens: it has already been copied into the intervening question and option tokens, and the answer position reads it from them.

Attention knockout remained effective, but only when applied throughout free-form generation rather than solely at the final answer position. In the multiple-choice setting, the model generates an answer from a small number of fixed positions, whereas free-form generation creates a sequence of new tokens, each of which can attend to information from earlier positions. We therefore tested which positions needed to be prevented from attending to the conflicting textual claim (Table~\ref{tab:routing}). Blocking only the later positions where the answer would be produced had little effect: emotional-prosody accuracy remained unchanged (0.000 to 0.000), while speaker-identity accuracy increased modestly (0.067 to 0.083). In contrast, blocking every position after the claim from attending to it substantially reduced textual dominance, increasing audio-consistent accuracy from 0.000 to 0.533 for emotional prosody and from 0.067 to 0.717 for speaker identity. Under this intervention, TDR fell from 0.983 to 0.150 for prosody and from 0.933 to 0.283 for identity, at a capability cost of 1.000 to 0.367 and 1.000 to 0.800 respectively. These results indicate that the relevant distinction is not which layers are intervened on, but which token positions are blocked: in the multiple-choice setting, only the final answer position is affected, whereas free-form generation allows the intervention to block the entire sequence of tokens that could otherwise carry the textual claim forward. The results therefore suggest that the model's textual bias is mediated by an intermediate relay pathway, in which the claim is repeatedly re-encoded into later tokens before influencing the final response.

\begin{table}[!htbp]
\centering
\caption{Routing bisection: audio-consistent accuracy under closed-set scoring
when different token positions are blocked from attending to the claim (all
layers, $\gamma{=}0$). Blocking the answer position alone recovers none of the
effect; blocking every position downstream of the claim recovers all of it.}
\label{tab:routing}
\small
\begin{tabular}{lcccccc}
\toprule
& \multicolumn{3}{c}{Emotional Prosody} & \multicolumn{3}{c}{Speaker Identity} \\
\cmidrule(lr){2-4}\cmidrule(lr){5-7}
Positions blocked & Audio acc. & $\Delta$ margin & Capability & Audio acc. & $\Delta$ margin & Capability \\
\midrule
None (baseline)            & 0.000 & ---      & 1.000 & 0.067 & ---      & 1.000 \\
Answer position only       & 0.000 & $-0.00$  & 1.000 & 0.083 & $+0.01$  & 1.000 \\
All positions after claim  & \textbf{0.533} & $+10.54$ & 0.367 & \textbf{0.717} & $+5.80$ & 0.800 \\
Every position             & 0.533 & $+10.54$ & 0.367 & 0.717 & $+5.80$ & 0.800 \\
\bottomrule
\end{tabular}
\end{table}

\subsection{Controls}

We ran controls to confirm that the intervention was not simply damaging model performance in a way that mechanically inflated TDR, and that similar interventions inconsistent with our account do not work. On matched items, accuracy increased from 0.767 to 0.900 for prosody and from 0.933 to 1.000 for identity. The model demonstrated identical accuracy for prosody and speaker identity when given text alone with no audio. This effect appeared only when blocking mid-layers; blocking late layers did not affect TDR for either category. Finally, blocking at all layers suppressed TDR further but impaired the model's ability, dropping matched-case prosody accuracy to 0.667. This rules out the possibility that the intervention simply deletes the text claim from the input and forces the model to rely on audio by default. Full per-condition results are given in Table~\ref{tab:freeform}.

\begin{table}[!htbp]
\centering
\caption{Free-form generation under attention knockout. gen-TDR is the fraction
of descriptions naming the false claimed label (lower is better); ``names true''
is the fraction naming the audio-supported label. Capability is accuracy on
matched items; cost is accuracy on text-only items, where no audio is present
and the claim is the only evidence. Fluency is mean per-token log-probability
under the unmodified model, scored in a claim-free context. 95\% Wilson
intervals.}
\label{tab:freeform}
\small
\begin{tabular}{llccccc}
\toprule
Category & Condition & gen-TDR [95\% CI] & Names true & Capability & Cost & Fluency \\
\midrule
\multirow{4}{*}{Prosody}
 & Baseline           & 1.000 [0.86, 1.00] & 0.000 & 0.767 & 0.567 & $-1.28$ \\
 & Late $\gamma{=}0$  & 1.000 [0.86, 1.00] & 0.000 & 0.900 & 0.467 & $-1.13$ \\
 & \textbf{Mid $\gamma{=}0$} & \textbf{0.375} [0.21, 0.57] & \textbf{0.500} & 0.900 & 0.567 & $-0.64$ \\
 & All $\gamma{=}0$   & 0.182 [0.07, 0.39] & 0.600 & 0.667 & 0.533 & $-0.46$ \\
\midrule
\multirow{4}{*}{Identity}
 & Baseline           & 1.000 [0.89, 1.00] & 0.000 & 0.933 & 1.000 & $-1.45$ \\
 & Late $\gamma{=}0$  & 1.000 [0.87, 1.00] & 0.033 & 1.000 & 1.000 & $-1.54$ \\
 & \textbf{Mid $\gamma{=}0$} & \textbf{0.433} [0.27, 0.61] & \textbf{0.567} & 1.000 & 1.000 & $-0.78$ \\
 & All $\gamma{=}0$   & 0.300 [0.17, 0.48] & 0.700 & 0.933 & 0.900 & $-0.69$ \\
\bottomrule
\end{tabular}
\end{table}

\section{Discussion}
\label{sec:discussion}

Our repair evidence divides into what back-patching can and cannot tell us.
Because it produced no measurable effect even on the lexical task it was
designed for (Table~\ref{tab:alme}), its null on paralinguistic conflict carries
no information about whether the two phenomena share a mechanism, and we draw no
such conclusion. The audio-side evidence that does bear weight comes from probing
and amplification. This is not an availability failure: probes recover speaker
gender from the encoder at ceiling from layer 13 onward and the information
remains linearly decodable at the projector output. The acoustic evidence reaches
the language model and is ignored. Amplification points the same way, where
scaling the audio span 5x left TDR unchanged while capability collapsed from
1.000 to 0.100, so interventions that raise audio salience are targeting the
wrong variable. 

The two attention-knockout conditions locate where the asymmetry lives. Blocking
the answer position from attending to the claim leaves TDR unchanged, which
indicates the claim has already been incorporated into the question and option
representations in the early layers. Blocking every generated position across
mid-layers during free-form generation moves it substantially. We conclude that
the claim's influence is distributed across intermediate positions and
reconstructed at each generation step, rather than passed through a single site
that one patch can sever. Concurrent work applying causal tracing to audio--text
fusion reports a compatible structure in DeSTA, Qwen and Voxtral: the final
sequence position acts as an informational bottleneck, and intermediate
positions perform an attention-like retrieval of task-relevant audio context
\citep{chen2026causaltracing}. That analysis is descriptive and does not involve
conflicting inputs, so it does not test suppression; but it independently
locates the two components our routing account depends on.

Two alternative explanations should be addressed. First, the mid-layer knockout
might degrade processing generally rather than remove the claim specifically.
Our controls rule this out by showing that matched-item accuracy under the intervention rises rather than falls in both categories. Second, the gap between multiple-choice
and free-form results is larger than the gap between blocking patterns, so the
evaluation format may itself be driving the effect. These are not independent, and
our account predicts exactly this: option tokens multiply the positions carrying
claim-informed content, so removing any one leaves others to propagate it.
Evaluation format is therefore not incidental to measuring text dominance; it
determines how much redundancy the mechanism has to work with.

\section{Limitations}

Our datasets use acted performances from RAVDESS \citep{livingstone2018ravdess} and MUStARD++ \citep{ray2022mustardpp}, so their paralinguistic signals and audio-text conflicts may be more categorical than in natural conversation. All conflicts are also in English, leaving open whether the observed mechanisms generalize across languages, particularly where prior work reports different conflict rates \citep{billa2026alme}. We evaluate emotional prosody, speaker identity, and sarcasm/pacing, but the mechanistic analysis excludes sarcasm/pacing because the audio-only labels were insufficiently determined to support a suppression claim. Other forms of paralinguistic conflict, including non-speech events, accent/dialect, and vocal effort, remain untested.

Our mechanistic analysis and repair are primarily demonstrated in a single model and on two conflict categories. The second model establishes the prerequisite that paralinguistic information is recoverable, but does not independently validate the identified mechanism. The intervention also has important scope conditions: it reduces text dominance during free-form generation but not closed-set multiple-choice scoring, and its effect depends on attenuating the textual claim across the intermediate positions that can relay its information. Our lexical back-patching replication produced a null despite the procedure's reported effectiveness; because our task and TTS instantiation differ from the original study, we interpret this cautiously as evidence of configuration sensitivity rather than a contradiction.

Finally, the intervention is unconditional: it attenuates the claim whether or not the claim is correct. Although text-only performance is preserved, we do not test cases where ambiguous audio should appropriately be overridden by a reliable textual claim. The free-form intervention experiments also use a limited number of items, so individual configuration differences should not be over-interpreted. More broadly, the same sensitivity to paralinguistic information that enables audio-grounded reasoning can facilitate inference of sensitive attributes such as emotion or identity, raising privacy and consent concerns for deployment.

\section{Conclusion}

We investigated whether paralinguistic suppression in audio LLMs is mechanistically related to lexical text dominance and whether it can be repaired by the interventions developed for that setting. Across six models, text-dominance under conflict is widespread and statistically robust (Cochran's Q, all categories p < 1e-40). In the model with the strongest and most consistently measurable bias, probing showed that paralinguistic information survives to the language model's input in full (ceiling accuracy at the projector), so the failure is not one of perception. Direct audio amplification failed to restore audio-grounded answers at any scale tested, ruling out an availability-based account; back-patching \citep{cho2026conflict} produced no measurable effect in either setting, including our replication on the lexical task it was designed for, and therefore constrains no hypothesis. A routing analysis located the mechanism: the claim's influence is not carried through a single connection to the answer position but is re-encoded into intervening tokens, which is why blocking that direct connection has no effect. Targeting the relay instead by attenuating the claim across mid-layers during free-form generation reduced text-dominance substantially in both categories tested, with matched-item accuracy and text-only performance unaffected. The effect appears only in open-ended generation but not in the closed-set multiple-choice format common in audio LLM evaluation, suggesting that benchmarks relying on that format may be systematically blind to interventions of this kind.

\section*{Reproducibility Statement}
To facilitate the reproducibility of our mechanistic diagnosis pipeline, detailed prompt templates and data construction methodology for the 3,594-item paralinguistic conflict benchmark—derived from the MUSTARD++ and RAVDESS corpora—are provided in Appendix A.6. This includes comprehensive examples of prompt formatting across all conditions for emotional prosody, speaker identity, and sarcasm/pacing conflicts. The complete evaluation codebase, including all scripts required for linear probing, attribution patching, and attention knockout, along with the full dataset manifests and experimental hyperparameters, will be made publicly available in an open-source repository upon publication.

\bibliographystyle{plainnat}
\bibliography{references}

@misc{abouelenin2025phi4,
  title         = {{Phi-4-mini technical report: Compact yet powerful multimodal language models via mixture-of-LoRAs}},
  author        = {A. Abouelenin and A. Ashfaq and A. Atkinson and H. Awadalla and N. Bach and J. Bao and A. Benhaim and M. Cai and V. Chaudhary and C. Chen and others},
  eprint        = {2503.01743},
  archivePrefix = {arXiv},
  howpublished  = {arXiv preprint arXiv:2503.01743},
  year          = {2025},
}

@misc{chu2024qwen2audio,
  title         = {{Qwen2-Audio technical report}},
  author        = {Y. Chu and J. Xu and Q. Yang and H. Wei and X. Wei and Z. Guo and Y. Leng and Y. Lv and J. He and J. Lin and others},
  eprint        = {2407.10759},
  archivePrefix = {arXiv},
  howpublished  = {arXiv preprint arXiv:2407.10759},
  year          = {2024},
}

@misc{li2025baichuan,
  title         = {{Baichuan-Omni-1.5 technical report}},
  author        = {Y. Li and others},
  eprint        = {2501.15368},
  archivePrefix = {arXiv},
  howpublished  = {arXiv preprint arXiv:2501.15368},
  year          = {2025},
}

@misc{lu2025desta25,
  title         = {{DeSTA2.5-Audio: Toward general-purpose large audio language model with self-generated cross-modal alignment}},
  author        = {K.-H. Lu and Z. Chen and S.-W. Fu and C.-H. H. Yang and J. Balam and B. Ginsburg and Y.-C. F. Wang and H.-Y. Lee},
  eprint        = {2507.02768},
  archivePrefix = {arXiv},
  howpublished  = {arXiv preprint arXiv:2507.02768},
  year          = {2025},
}

@misc{nvidia2026audex,
  title         = {{Unified audio intelligence without regressing on text intelligence}},
  author        = {NVIDIA Nemotron Labs},
  eprint        = {2607.05196},
  archivePrefix = {arXiv},
  howpublished  = {arXiv preprint arXiv:2607.05196},
  year          = {2026},
}

@inproceedings{radford2023whisper,
  title         = {{Robust speech recognition via large-scale weak supervision}},
  author        = {A. Radford and J. W. Kim and T. Xu and G. Brockman and C. McLeavey and I. Sutskever},
  booktitle     = {{International Conference on Machine Learning (ICML)}},
  year          = {2023},
}

@inproceedings{tang2024salmonn,
  title         = {{SALMONN: Towards generic hearing abilities for large language models}},
  author        = {C. Tang and W. Yu and G. Sun and X. Chen and T. Tan and W. Li and L. Lu and Z. Ma and C. Zhang},
  booktitle     = {{International Conference on Learning Representations (ICLR)}},
  year          = {2024},
}

@misc{qwen2025qwen25,
  title         = {{Qwen2.5 technical report}},
  author        = {A. Yang and B. Yang and B. Zhang and B. Hui and B. Zheng and B. Yu and C. Li and D. Liu and F. Huang and H. Wei and others},
  eprint        = {2412.15115},
  archivePrefix = {arXiv},
  howpublished  = {arXiv preprint arXiv:2412.15115},
  year          = {2024},
}

@article{livingstone2018ravdess,
  title         = {{The Ryerson Audio-Visual Database of Emotional Speech and Song (RAVDESS): A dynamic, multimodal set of facial and vocal expressions in North American English}},
  author        = {S. R. Livingstone and F. A. Russo},
  journal       = {{PLoS ONE}},
  volume        = {13},
  number        = {5},
  pages         = {e0196391},
  year          = {2018},
}

@inproceedings{ray2022mustardpp,
  title         = {{A multimodal corpus for emotion recognition in sarcasm}},
  author        = {A. Ray and S. Mishra and A. Nunna and P. Bhattacharyya},
  booktitle     = {{Language Resources and Evaluation Conference (LREC)}},
  year          = {2022},
}

@misc{chen2026listen,
  title         = {{Do audio LLMs really LISTEN, or just transcribe? Measuring lexical vs. acoustic emotion cues reliance}},
  author        = {J.  Chen and Z.  Guo and J.  Chun and P.  Wang and A.  Perrault and M.  Elsner},
  eprint        = {2510.10444},
  archivePrefix = {arXiv},
  howpublished  = {arXiv preprint arXiv:2510.10444},
  year          = {2025},
}

@misc{cho2026conflict,
  title         = {{Who wins the conflict? Mechanistic interpretability of text bias in audio LLMs}},
  author        = {H.  Cho and S.  Yoo and J.  Jang and C.  Kim and J.  Chung},
  eprint        = {2606.18924},
  archivePrefix = {arXiv},
  howpublished  = {arXiv preprint arXiv:2606.18924},
  year          = {2026},
}

@misc{billa2026alme,
  title         = {{When audio-LLMs don't listen: A cross-linguistic study of modality arbitration (ALME benchmark)}},
  author        = {Billa},
  howpublished  = {2026},
  year          = {2026},
}

@misc{mcrbench,
  title         = {{When audio and text disagree: Benchmarking text bias in large audio-language models under cross-modal inconsistencies}},
  author        = {C. Wang and G. Deng and X. Yang and H. Qiu and T. Zhang},
  eprint        = {2508.15407},
  archivePrefix = {arXiv},
  howpublished  = {arXiv preprint arXiv:2508.15407},
  year          = {2025},
}

@misc{alain2016probing,
  title         = {{Understanding intermediate layers using linear classifier probes}},
  author        = {G. Alain and Y. Bengio},
  eprint        = {1610.01644},
  archivePrefix = {arXiv},
  howpublished  = {arXiv preprint arXiv:1610.01644},
  year          = {2016},
}

@article{belinkov2022probing,
  title         = {{Probing classifiers: Promises, shortcomings, and advances}},
  author        = {Y. Belinkov},
  journal       = {{Computational Linguistics}},
  volume        = {48},
  number        = {1},
  pages         = {207--219},
  year          = {2022},
}

@inproceedings{geva2023dissecting,
  title         = {{Dissecting recall of factual associations in auto-regressive language models}},
  author        = {M. Geva and J. Bastings and K. Filippova and A. Globerson},
  booktitle     = {{EMNLP}},
  year          = {2023},
}

@misc{hanna2024faithfulness,
  title         = {{Have faith in faithfulness: Going beyond circuit overlap when finding model mechanisms}},
  author        = {M. Hanna and S. Pezzelle and Y. Belinkov},
  eprint        = {2403.17806},
  archivePrefix = {arXiv},
  howpublished  = {arXiv preprint arXiv:2403.17806},
  year          = {2024},
}

@inproceedings{meng2022rome,
  title         = {{Locating and editing factual associations in GPT}},
  author        = {K. Meng and D. Bau and A. Andonian and Y. Belinkov},
  booktitle     = {{Advances in Neural Information Processing Systems (NeurIPS)}},
  year          = {2022},
}

@article{syed2023attribution,
  title         = {{Attribution patching outperforms automated circuit discovery}},
  author        = {A. Syed and C. Rager and A. Conmy},
  journal       = {{arXiv preprint}},
  year          = {2023},
}

@inproceedings{vig2020causal,
  title         = {{Investigating gender bias in language models using causal mediation analysis}},
  author        = {J. Vig and S. Gehrmann and Y. Belinkov and S. Qian and D. Nevo and Y. Singer and S. Shieber},
  booktitle     = {{NeurIPS}},
  year          = {2020},
}

@inproceedings{wang2023ioi,
  title         = {{Interpretability in the wild: A circuit for indirect object identification in GPT-2 small}},
  author        = {K. Wang and A. Variengien and A. Conmy and B. Shlegeris and J. Steinhardt},
  booktitle     = {{ICLR}},
  year          = {2023},
}

@inproceedings{zhang2024pasta,
  title         = {{Tell your model where to attend: Post-hoc attention steering for LLMs}},
  author        = {Q. Zhang and C. Singh and L. Liu and X. Liu and B. Yu and J. Gao and T. Zhao},
  booktitle     = {{ICLR}},
  year          = {2024},
}

@inproceedings{li2023iti,
  title         = {{Inference-time intervention: Eliciting truthful answers from a language model}},
  author        = {K. Li and O. Patel and F. Vi{\'e}gas and H. Pfister and M. Wattenberg},
  booktitle     = {{NeurIPS}},
  year          = {2023},
}

@inproceedings{shi2024cad,
  title         = {{Trusting your evidence: Hallucinate less with context-aware decoding}},
  author        = {W. Shi and X. Han and M. Lewis and Y. Tsvetkov and L. Zettlemoyer and S. W. Yih},
  booktitle     = {{NAACL}},
  year          = {2024},
}

@inproceedings{longpre2021entity,
  title         = {{Entity-based knowledge conflicts in question answering}},
  author        = {S. Longpre and K. Perisetla and A. Chen and N. Ramesh and C. DuBois and S. Singh},
  booktitle     = {{EMNLP}},
  year          = {2021},
}

@inproceedings{xu2024knowledgeconflict,
  title         = {{Knowledge conflicts for LLMs: A survey}},
  author        = {R. Xu and Z. Qi and Z. Guo and C. Wang and H. Wang and Y. Zhang and W. Xu},
  booktitle     = {{EMNLP}},
  year          = {2024},
}

@article{cochran1950,
  title         = {{The comparison of percentages in matched samples}},
  author        = {W. G. Cochran},
  journal       = {{Biometrika}},
  year          = {1950},
}

@article{holm1979,
  title         = {{A simple sequentially rejective multiple test procedure}},
  author        = {S. Holm},
  journal       = {{Scandinavian Journal of Statistics}},
  volume        = {6},
  number        = {2},
  pages         = {65--70},
  year          = {1979},
}

@article{mcnemar1947,
  title         = {{Note on the sampling error of the difference between correlated proportions or percentages}},
  author        = {Q. McNemar},
  journal       = {{Psychometrika}},
  volume        = {12},
  number        = {2},
  pages         = {153--157},
  year          = {1947},
}

@article{cummins2015review,
  title         = {{A review of depression and suicide risk assessment using speech analysis}},
  author        = {N. Cummins and S. Scherer and J. Krajewski and S. Schnieder and J. Epps and T. F. Quatieri},
  journal       = {{Speech Communication}},
  year          = {2015},
}

@article{hansen2015speaker,
  title         = {{Speaker recognition by machines and humans: A tutorial review}},
  author        = {J. H. L. Hansen and T. Hasan},
  journal       = {{IEEE Signal Processing Magazine}},
  volume        = {32},
  number        = {6},
  pages         = {74--99},
  year          = {2015},
}

@article{schuller2013paralinguistics,
  title         = {{Paralinguistics in speech and language---state-of-the-art and the challenge}},
  author        = {B. Schuller and S. Steidl and A. Batliner and F. Burkhardt and L. Devillers and C. M{\"u}ller and S. Narayanan},
  journal       = {{Computer Speech \& Language}},
  volume        = {27},
  number        = {1},
  pages         = {4--39},
  year          = {2013},
}

@misc{parsabench,
  title         = {{PARSA-Bench: A Comprehensive Persian Audio-Language Model Benchmark}},
  author        = {Mohammad Javad Ranjbar Kalahroodi and Mohammad Amini and Parmis Bathayan and Heshaam Faili and Azadeh Shakery},
  eprint        = {2603.14456},
  archivePrefix = {arXiv},
  howpublished  = {arXiv preprint arXiv:2603.14456},
  year          = {2026},
}

@misc{chen2026causaltracing,
  title         = {{Causal Tracing of Audio-Text Fusion in Large Audio Language Models}},
  author        = {Wei-Chih Chen and Huang, Chien-yu and Lee, Hung-yi},
  eprint        = {2603.13768},
  archivePrefix = {arXiv},
  howpublished  = {arXiv preprint arXiv:2603.13768},
  year          = {2026},
}

\appendix
\raggedbottom
\makeatletter
\setlength{\@fptop}{0pt}
\setlength{\@fpsep}{12pt}
\setlength{\@fpbot}{0pt}
\makeatother

\section{Appendix}
\label{app:pairwise}

\subsection{Extended Related Work}
\label{app:related}

\paragraph{Inference-time attention steering.}
A line of work resolves conflicts between competing sources of information without
retraining. PASTA \citep{zhang2024pasta} identifies a small set of attention heads and
reweights them to emphasise user-specified text spans, improving performance on tasks
involving conflicting information within a text context. Other methods steer models toward
contextual or parametric knowledge by shifting activations along directions identified
from probing \citep{li2023iti} or by contrastive decoding that upweights
context-conditioned predictions \citep{shi2024cad}. The broader problem of conflict
between contextual and parametric knowledge has been studied extensively in text-only
settings \citep{longpre2021entity, xu2024knowledgeconflict}. Together these show that
modifying attention at inference time can influence which source a language model
prioritises, but all concern conflicts entirely within text: the competing information
arrives through the same textual stream, whether from different spans, documents, or the
model's own parameters.

Our setting differs in that one source is acoustic and must pass through an audio encoder
and projector before entering the residual stream, where it competes with text. It is
therefore not obvious that steering methods designed for text conflicts transfer. We find
that the direct adaptation --- blocking the answer position from attending to the
conflicting claim --- is insufficient, because the claim is re-encoded into intermediate
tokens that carry it forward. Reducing textual dominance requires intervening on this
relay rather than at the claim's source or at the answer position.

\paragraph{Behavioural benchmarks for audio--text conflict.}
MCR-Bench \citep{mcrbench} measures whether audio LLMs follow adversarial textual
descriptions over their own perception on sound and vocal classification tasks.
PARSA-Bench \citep{parsabench}, a Persian-language audio benchmark, reports that text-only
baselines consistently outperform their audio counterparts, indicating that reliance on
transcript content over acoustic evidence is not confined to English. LISTEN
\citep{chen2026listen} separates lexical and acoustic information through neutral,
matched, mismatched, and paralinguistic conditions, finding consistent lexical dominance
across six models with accuracy improving only slightly when cues align. All three
establish behavioural rates without localising the components responsible.

\subsection{Per-Model Text Dominance Rate}
\label{app:tdr-table}

Table~\ref{tab:app-tdr} reports the complete-case Text Dominance Rate (TDR)
for all six models across the three conflict categories, supporting the
TDR claims summarized in the main Results section. Complete-case TDR
excludes trials with invalid or refused model outputs (see
the main Results section for exclusion criteria); per-category dropped-trial
counts are given in Table~\ref{tab:app-dropped}.

\begin{table}[!htbp]
\centering
\caption{Complete-case Text Dominance Rate (\%) by model and conflict category.}
\label{tab:app-tdr}
\small
\begin{tabular}{lcccccc}
\toprule
Category & Audex & Baichuan & DeSTA2.5 & Phi-4-mm & Qwen2-Audio & SALMONN \\
\midrule
Emotional Prosody & 98.7 & 78.3 & 68.8 & 93.1 & 98.7 & 83.9 \\
Speaker Identity   & 96.4 & 89.1 & 98.5 & 96.1 & 94.9 & 70.7 \\
Sarcasm/Pacing     & 34.4 & 24.7 & 20.9 & 65.3 & 58.5 & 56.1 \\
\bottomrule
\end{tabular}
\end{table}

\begin{table}[!htbp]
\centering
\caption{Complete-case sample sizes and dropped trials (invalid/refused outputs) by category.}
\label{tab:app-dropped}
\small
\begin{tabular}{lccc p{4.2cm}}
\toprule
Category & Total Trials & Complete-case $n$ & Dropped & Driven by \\
\midrule
Emotional Prosody & 399 & 304 & 95  & Baichuan (86), DeSTA2.5 (11), Qwen2-Audio (2) \\
Speaker Identity   & 450 & 331 & 119 & Baichuan (119) \\
Sarcasm/Pacing     & 399 & 369 & 30  & Qwen2-Audio (15), Baichuan (12), DeSTA2.5 (4) \\
\bottomrule
\end{tabular}
\end{table}

\subsection{Per-Model Accuracy by Condition}
\label{app:accuracy-tables}

Tables~\ref{tab:app-acc-prosody}--\ref{tab:app-acc-sarcasm} report accuracy
(\%) for all six models across the four evaluation conditions
(Neutral-Text, Category-Matched, Category-Mismatched, Paralinguistic-only),
for each conflict category.

\begin{table}[!htbp]
\centering
\caption{Emotional Prosody: accuracy (\%) by condition and model.}
\label{tab:app-acc-prosody}
\small
\begin{tabular}{lcccccc}
\toprule
Condition & Audex & Baichuan & DeSTA2.5 & Phi-4-mm & Qwen2-Audio & SALMONN \\
\midrule
Neutral-Text        & 75.9 & 44.2 & 55.5 & 52.3 & 49.8 & 16.5 \\
Category-Matched    & 99.2 & 78.3 & 86.9 & 93.2 & 100.0 & 90.2 \\
Category-Mismatched & 34.1 & 43.8 & 52.1 & 32.3 & 33.0 & 29.3 \\
Paralinguistic-only & 60.9 & 9.9  & 50.4 & 13.5 & 54.1 & 15.0 \\
\bottomrule
\end{tabular}
\end{table}

\begin{table}[!htbp]
\centering
\caption{Speaker Identity: accuracy (\%) by condition and model.}
\label{tab:app-acc-identity}
\small
\begin{tabular}{lcccccc}
\toprule
Condition & Audex & Baichuan & DeSTA2.5 & Phi-4-mm & Qwen2-Audio & SALMONN \\
\midrule
Neutral-Text        & 92.7 & 47.1 & 99.2  & 57.3 & 98.6  & 58.0 \\
Category-Matched    & 99.3 & 94.1 & 100.0 & 96.9 & 100.0 & 65.8 \\
Category-Mismatched & 37.3 & 54.7 & 34.7  & 37.3 & 38.4  & 41.6 \\
Paralinguistic-only & 92.0 & 56.1 & 99.3  & 51.3 & 98.7  & 44.7 \\
\bottomrule
\end{tabular}
\end{table}

\begin{table}[!htbp]
\centering
\caption{Sarcasm/Pacing: accuracy (\%) by condition and model.}
\label{tab:app-acc-sarcasm}
\small
\begin{tabular}{lcccccc}
\toprule
Condition & Audex & Baichuan & DeSTA2.5 & Phi-4-mm & Qwen2-Audio & SALMONN \\
\midrule
Neutral-Text        & 20.3 & 20.0 & 21.4 & 23.3 & 16.9 & 13.9 \\
Category-Matched    & 46.1 & 29.6 & 21.3 & 70.9 & 68.5 & 56.6 \\
Category-Mismatched & 14.0 & 14.2 & 12.4 & 9.5  & 9.1  & 7.3 \\
Paralinguistic-only & 19.5 & 13.8 & 21.1 & 13.5 & 16.9 & 13.5 \\
\bottomrule
\end{tabular}
\end{table}

\subsection{Audio-Side Intervention Sweeps}
\label{app:exp5}

Table~\ref{tab:app-exp5} reports the source-layer, destination-window, and
amplitude sweeps for the audio-side interventions described in the main
Results section, on emotional prosody with untruncated audio.

\begin{table}[!htbp]
\centering
\caption{Audio-side intervention sweeps (emotional prosody, untruncated audio).
Decision margin is $\log P(\text{audio label}) - \log P(\text{claimed label})$
in nats; baseline margin is $-7.407$. No configuration closes more than 12\% of
the gap.}
\label{tab:app-exp5}
\small
\begin{tabular}{lcccccccc}
\toprule
\multicolumn{9}{l}{\textit{(a) Source layer sweep} ($w{=}1$, $\mathrm{dst}{=}0$)} \\
\midrule
Source layer & 4 & 9 & 13 & 16 & 20 & 24 & 28 & 31 \\
$\Delta$ margin (nats) & $+0.17$ & $+0.42$ & $+0.66$ & $\mathbf{+0.87}$ & $+0.47$ & $+0.04$ & $-0.11$ & $-0.19$ \\
\midrule
\multicolumn{9}{l}{\textit{(b) Destination window sweep} (src${=}13$)} \\
\midrule
Window width & \multicolumn{2}{c}{1} & \multicolumn{2}{c}{3} & \multicolumn{2}{c}{5} & \multicolumn{2}{c}{7} \\
$\Delta$ margin (nats) & \multicolumn{2}{c}{$+0.66$} & \multicolumn{2}{c}{$+0.60$} & \multicolumn{2}{c}{$+0.58$} & \multicolumn{2}{c}{$+0.59$} \\
\midrule
\multicolumn{9}{l}{\textit{(c) Amplitude sweep} (audio span scaled in place)} \\
\midrule
Scale factor & \multicolumn{2}{c}{1.5$\times$} & \multicolumn{2}{c}{2.0$\times$} & \multicolumn{2}{c}{3.0$\times$} & \multicolumn{2}{c}{5.0$\times$} \\
Capability & \multicolumn{2}{c}{1.000} & \multicolumn{2}{c}{1.000} & \multicolumn{2}{c}{1.000} & \multicolumn{2}{c}{0.100} \\
\bottomrule
\end{tabular}
\end{table}

\subsection{Layer-Index Convention Check}
\label{app:indexing}

\begin{table}[!htbp]
\centering
\caption{Layer-index convention check. The published configurations are given
without a stated indexing base; re-running all four under the alternative
reading (both endpoints shifted down one) leaves the null unchanged.}
\label{tab:app-indexing}
\small
\begin{tabular}{lcc}
\toprule
Flip type & 0-indexed gain & 1-indexed gain \\
\midrule
Adjective swap & $+0.013$ & $+0.013$ \\
Number swap    & $+0.000$ & $+0.000$ \\
Time swap      & $+0.040$ & $+0.013$ \\
Negation add   & $-0.027$ & $-0.013$ \\
\midrule
\textbf{Pooled} ($n{=}300$) & $\mathbf{+0.007}$ & $\mathbf{+0.003}$ \\
Paired 95\% CI & $[-0.027, +0.040]$ & $[-0.028, +0.035]$ \\
McNemar exact $p$ & 0.845 & 1.000 \\
\bottomrule
\end{tabular}
\end{table}

\subsection{Dataset Construction Examples}
\label{app:dataset-examples}

Tables~\ref{tab:app-ex-prosody}--\ref{tab:app-ex-sarcasm} show example
prompts for each of the four conditions (Neutral-Text, Category-Matched,
Category-Mismatched, Paralinguistic-only), for each of the three conflict
categories used in the final dataset.

\begin{table}[!htbp]
\centering
\small
\caption{Example prompts: Emotional Prosody Conflict (audio emotion = happy).}
\label{tab:app-ex-prosody}
\begin{tabular}{p{2.4cm}p{9.5cm}}
\toprule
Condition & Example \\
\midrule
Neutral-Text & Transcript is ``The kids are sitting by the door.'' How would you describe the emotional tone of the speaker? \\
Category-Matched & The speaker's emotion is happy. Listen to the audio and classify the emotion. How would you describe the emotional tone of the speaker? Beware that the audio description and audio itself may not match. \\
Category-Mismatched & The speaker's emotion is angry. Listen to the audio and classify the emotion. How would you describe the emotional tone of the speaker? Beware that the audio description and audio itself may not match. \\
Paralinguistic-only & Listen to the audio and classify the emotion. What emotion is expressed in the speaker's voice? \\
\bottomrule
\end{tabular}
\end{table}

\begin{table}[!htbp]
\centering
\small
\caption{Example prompts: Speaker Identity Conflict (speaker identity = adult female).}
\label{tab:app-ex-identity}
\begin{tabular}{p{2.4cm}p{9.5cm}}
\toprule
Condition & Example \\
\midrule
Neutral-Text & Transcript is ``The kids are sitting by the door.'' What is the speaker's gender identity? \\
Category-Matched & The speaker is an adult female. Listen to the audio and classify the gender identity. What is the speaker's gender identity? Beware that the audio description and audio itself may not match. \\
Category-Mismatched & The speaker is an adult male. Listen to the audio and classify the gender identity. What is the speaker's gender identity? Beware that the audio description and audio itself may not match. \\
Paralinguistic-only & Listen to the audio and classify the gender identity. What is the speaker's gender identity? \\
\bottomrule
\end{tabular}
\end{table}

\begin{table}[!htbp]
\centering
\small
\caption{Example prompts: Sarcasm/Pacing Conflict (speaker is being sincere).}
\label{tab:app-ex-sarcasm}
\begin{tabular}{p{2.4cm}p{9.5cm}}
\toprule
Condition & Example \\
\midrule
Neutral-Text & Transcript is ``The dog is playing outside.'' How sincere or satirical is the speaker being? \\
Category-Matched & The speaker is being sincere. Listen to the audio and classify how sincere or satirical the speaker is. Beware that the audio description and audio itself may not match. \\
Category-Mismatched & The speaker is being sarcastic. Listen to the audio and classify how sincere or satirical the speaker is. Beware that the audio description and audio itself may not match. \\
Paralinguistic-only & Listen to the audio and classify how sincere or sarcastic the speaker is. \\
\bottomrule
\end{tabular}
\end{table}

\subsection{Dataset and Model Licenses}
\label{app:licenses}

RAVDESS is released under CC BY-NC-SA 4.0 \citep{livingstone2018ravdess}; MUStARD++ is
released for academic research use with no formal license specified \citep{ray2022mustardpp}.
Of the evaluated models, Qwen2-Audio-7B-Instruct and SALMONN-13B are released under Apache 2.0,
Phi-4-multimodal-instruct under MIT, Baichuan-Omni-1.5 under Apache 2.0 with an accompanying
community license, and Audex under NVIDIA's one-way noncommercial license; DeSTA2.5-Audio's
model authors do not specify a license. All models and datasets were used strictly for
non-commercial academic research, consistent with these terms.

\subsection{Statistical Tests}
\label{app:stats}

Table~\ref{tab:app-cochran} reports the Cochran's Q omnibus test for each
conflict category (testing whether TDR differs across the six models),
followed by pairwise McNemar tests with Holm-Bonferroni correction
(Tables~\ref{tab:app-mcnemar-prosody}--\ref{tab:app-mcnemar-sarcasm}).

\begin{table}[!htbp]
\centering
\caption{Cochran's Q omnibus test for between-model TDR differences.}
\label{tab:app-cochran}
\begin{tabular}{lccccc}
\toprule
Category & $n$ (complete-case) & $Q$ & df & $p$-value & Significant ($p<.05$) \\
\midrule
Emotional Prosody & 304 & 201.04 & 5 & $1.702\times10^{-41}$ & Yes \\
Speaker Identity   & 331 & 218.09 & 5 & $3.815\times10^{-45}$ & Yes \\
Sarcasm/Pacing     & 369 & 345.00 & 5 & $2.082\times10^{-72}$ & Yes \\
\bottomrule
\end{tabular}
\end{table}

\begin{table}[!htbp]
\centering
\small
\caption{Emotional Prosody: pairwise McNemar test, Holm-Bonferroni corrected $p$-values.}
\label{tab:app-mcnemar-prosody}
\resizebox{\linewidth}{!}{%
\begin{tabular}{lcccccc}
\toprule
 & Audex & Baichuan & DeSTA2.5 & Phi-4-mm & Qwen2-Audio & SALMONN \\
 \midrule
 Audex        & --- & $7.79\times10^{-16}$ & $2.85\times10^{-25}$ & 0.0024 & 1.0000 & $1.97\times10^{-10}$ \\
 Baichuan     & $7.79\times10^{-16}$ & --- & 0.0115 & $8.11\times10^{-7}$ & $1.82\times10^{-14}$ & 0.1931 \\
 DeSTA2.5     & $2.85\times10^{-25}$ & 0.0115 & --- & $1.66\times10^{-15}$ & $3.22\times10^{-24}$ & 0.0002 \\
 Phi-4-mm     & 0.0024 & $8.11\times10^{-7}$ & $1.66\times10^{-15}$ & --- & 0.0024 & 0.0014 \\
 Qwen2-Audio  & 1.0000 & $1.82\times10^{-14}$ & $3.22\times10^{-24}$ & 0.0024 & --- & $6.35\times10^{-10}$ \\
 SALMONN      & $1.97\times10^{-10}$ & 0.1931 & 0.0002 & 0.0014 & $6.35\times10^{-10}$ & --- \\
 \bottomrule
 \end{tabular}}
 \end{table}

 \begin{table}[!htbp]
 \centering
 \small
 \caption{Speaker Identity: pairwise McNemar test, Holm-Bonferroni corrected $p$-values.}
 \label{tab:app-mcnemar-identity}
 \resizebox{\linewidth}{!}{%
\begin{tabular}{lcccccc}
 \toprule
  & Audex & Baichuan & DeSTA2.5 & Phi-4-mm & Qwen2-Audio & SALMONN \\
  \midrule
  Audex        & --- & 0.0035 & 0.4813 & 1.0000 & 1.0000 & $6.09\times10^{-19}$ \\
  Baichuan     & 0.0035 & --- & $3.35\times10^{-6}$ & 0.0112 & 0.0768 & $1.03\times10^{-6}$ \\
  DeSTA2.5     & 0.4813 & $3.35\times10^{-6}$ & --- & 0.4813 & 0.0768 & $9.68\times10^{-23}$ \\
  Phi-4-mm     & 1.0000 & 0.0112 & 0.4813 & --- & 1.0000 & $7.62\times10^{-20}$ \\
  Qwen2-Audio  & 1.0000 & 0.0768 & 0.0768 & 1.0000 & --- & $4.81\times10^{-22}$ \\
  SALMONN      & $6.09\times10^{-19}$ & $1.03\times10^{-6}$ & $9.68\times10^{-23}$ & $7.62\times10^{-20}$ & $4.81\times10^{-22}$ & --- \\
  \bottomrule
  \end{tabular}}
  \end{table}

  \begin{table}[!htbp]
  \centering
  \small
  \caption{Sarcasm/Pacing: pairwise McNemar test, Holm-Bonferroni corrected $p$-values.}
  \label{tab:app-mcnemar-sarcasm}
  \resizebox{\linewidth}{!}{%
\begin{tabular}{lcccccc}
  \toprule
   & Audex & Baichuan & DeSTA2.5 & Phi-4-mm & Qwen2-Audio & SALMONN \\
   \midrule
   Audex        & --- & 0.0074 & $2.53\times10^{-6}$ & $4.63\times10^{-24}$ & $2.42\times10^{-14}$ & $2.14\times10^{-9}$ \\
   Baichuan     & 0.0074 & --- & 0.3956 & $3.87\times10^{-30}$ & $7.58\times10^{-23}$ & $2.05\times10^{-17}$ \\
   DeSTA2.5     & $2.53\times10^{-6}$ & 0.3956 & --- & $2.33\times10^{-37}$ & $3.44\times10^{-31}$ & $3.78\times10^{-20}$ \\
   Phi-4-mm     & $4.63\times10^{-24}$ & $3.87\times10^{-30}$ & $2.33\times10^{-37}$ & --- & 0.0106 & 0.0044 \\
   Qwen2-Audio  & $2.42\times10^{-14}$ & $7.58\times10^{-23}$ & $3.44\times10^{-31}$ & 0.0106 & --- & 0.4558 \\
   SALMONN      & $2.14\times10^{-9}$ & $2.05\times10^{-17}$ & $3.78\times10^{-20}$ & 0.0044 & 0.4558 & --- \\
   \bottomrule
   \end{tabular}}
   \end{table}

\subsection{Supplementary Figures}
\label{app:figures}

\begin{figure}[!htbp]
  \centering
  \includegraphics[width=0.8\linewidth]{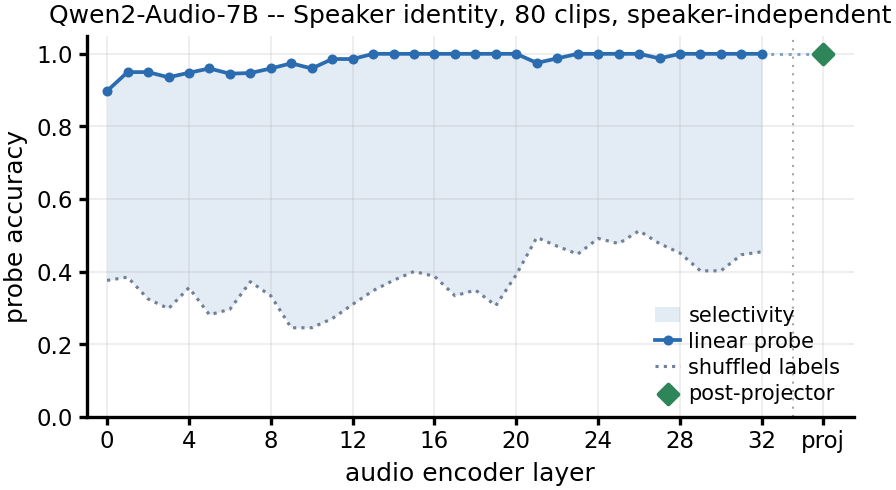}
  \caption{Linear-probe accuracy for speaker gender across encoder layers and at
  the projector output, with speaker-independent splits and a shuffled-label
  control.}
  \label{fig:app-probe}
\end{figure}

\begin{figure}[!htbp]
  \centering
  \includegraphics[width=0.8\linewidth]{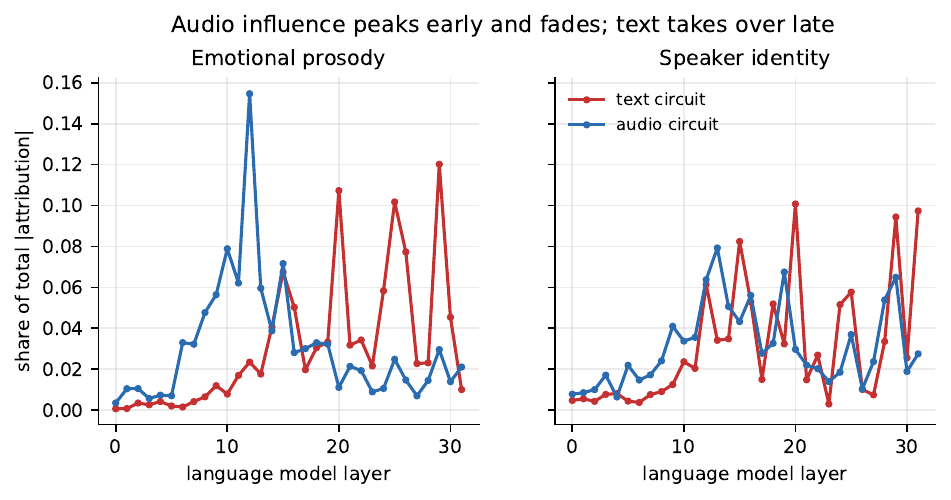}
  \caption{Full attribution map, including the audio position groups omitted from
  Figure~\ref{fig:layers-acc}. Text and audio panels share colour limits, so the
  near-uniform audio panels are the 18:1 text-to-audio ratio reported in
  Section~\ref{sec:retention} shown directly rather than summarised.}
  \label{fig:app-circuit}
\end{figure}

\begin{figure}[!htbp]
  \centering
  \includegraphics[width=0.8\linewidth]{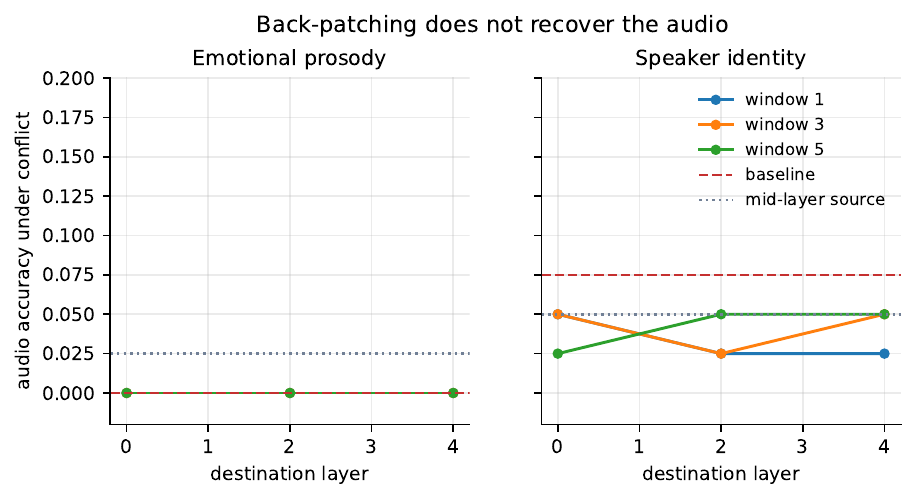}
  \caption{Back-patching sweep on paralinguistic conflicts across source layers,
  destination layers, and window widths.}
  \label{fig:app-backpatch}
\end{figure}

\begin{figure}[!htbp]
  \centering
  \includegraphics[width=\linewidth]{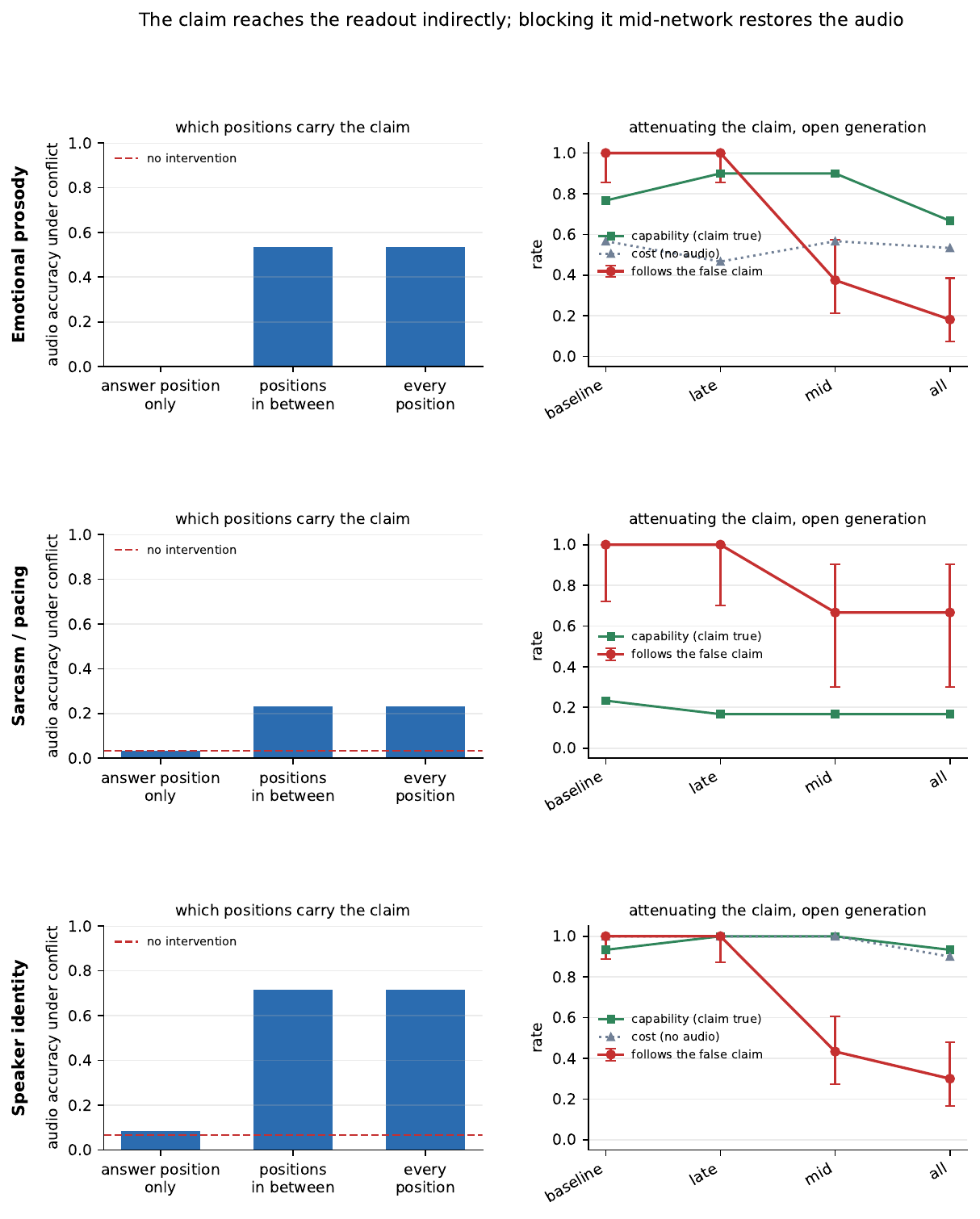}
  \caption{Left: routing bisection --- blocking the answer position's direct view
  of the claim recovers none of the effect, while blocking the intervening
  positions recovers all of it. Right: the generative intervention with 95\%
  Wilson intervals, alongside the capability and cost controls.}
  \label{fig:readout}
\end{figure}

\begin{figure}[!htbp]
  \centering
  \includegraphics[width=0.8\linewidth]{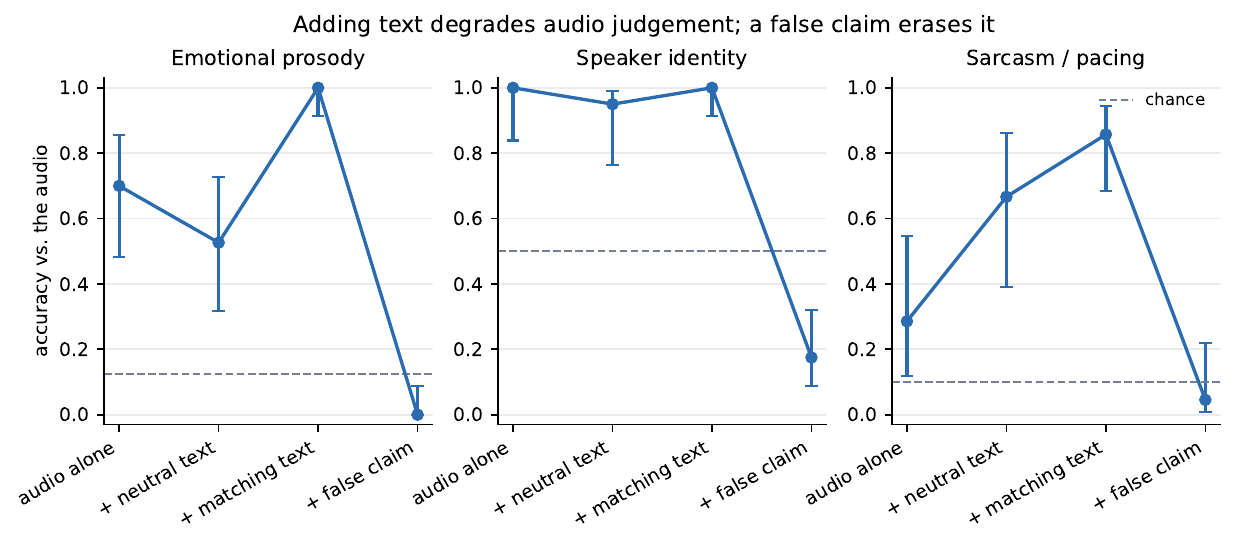}
  \caption{Accuracy across the four LISTEN conditions, by model and category.}
  \label{fig:app-ladder}
\end{figure}

\begin{figure}[!htbp]
  \centering
  \includegraphics[width=0.8\linewidth]{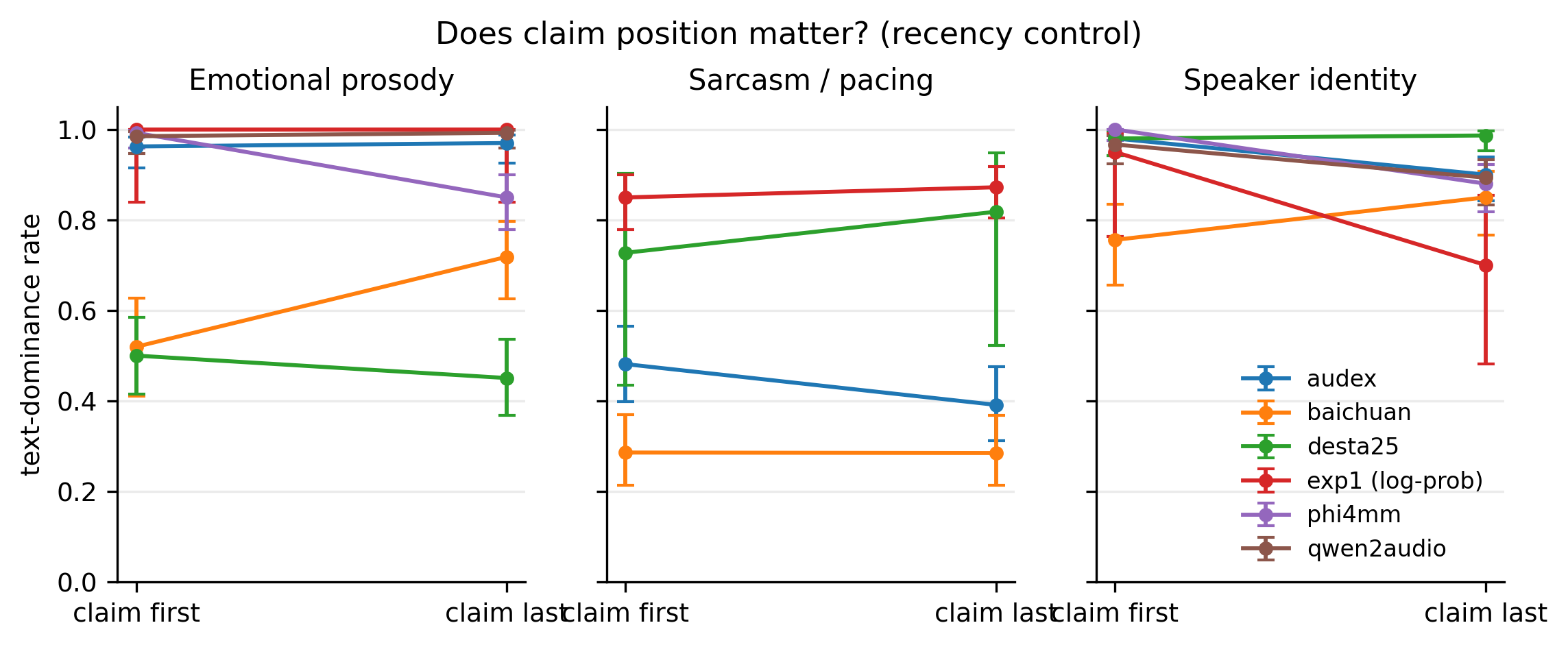}
  \caption{Text-dominance by claim order (claim-first vs.\ claim-last), as a control against recency bias. For emotional prosody and speaker identity, most models' text-dominance rate stays near ceiling regardless of claim position, indicating the effect is not simply an artifact of which claim appears more recently in the prompt. Sarcasm/pacing shows more movement between claim positions, but the direction differs by model, so claim order does not explain the overall text-dominance effect.}
  \label{fig:app-recency}

\end{figure}

\end{document}